\documentclass[a4paper,11pt]{article}

\usepackage{amsmath}
\usepackage{amssymb}
\usepackage{amsthm}
\usepackage{indentfirst}
\usepackage{graphicx}
\usepackage[utf8]{inputenc}
\usepackage[dvipsnames]{xcolor}
\usepackage{tabularx}
\usepackage{caption}
\usepackage{subcaption}
\usepackage{gensymb}
\usepackage[normalem]{ulem}
\usepackage{array}
\usepackage{enumitem}
\usepackage{makecell}
\usepackage{booktabs}

\usepackage{algorithm}
\usepackage{algpseudocode}

\usepackage{marginnote}

\usepackage{jinstpub}
\hypersetup{colorlinks=true, linkcolor=blue, urlcolor=blue, citecolor=blue, linktocpage=true}

\graphicspath{{images/}}

\begin{document}

\title{Safe Domain Adaptation for Physics: 
Overcoming Nuisances, Label Shifts, and Simulation Priors}

\author[a,b] {I.~Kharuk}

\affiliation[a]{Institute for Nuclear Research of the Russian Academy of Sciences, \\ 60th October Anniversary Prospect 7a, Moscow, 117312, Russia}
\affiliation[b]{Moscow Institute of Physics and Technology,\\ Institutsky lane, 9, Dolgoprudny, Moscow region, 141700, Russia}

\emailAdd{ivan.kharuk@phystech.edu}

\abstract{
Domain adaptation is widely used to make neural networks trained on simulations applicable to experimental data. 
Its premise is that the two domains differ only in nuisances, and that the quantity of interest is distributed identically in both. 
In physics neither assumption holds: simulations can be wrong about the physics, and the distribution of the target quantity — an energy spectrum, a redshift distribution — is often the measurement itself. 
We study the consequences of such mismatches on a toy air-shower benchmark in which a detector-response nuisance, a physical simulation shift, and an energy-spectrum shift can be switched on separately or together. 
Standard adversarial adaptation handles the conditional shifts, but once the two spectra differ it aligns them, replacing an uncontrolled bias by one anchored on the simulation prior.
We present adaptive domain adaptation, which reweights the simulated events so as to focus domain adaptation on the genuine physical mismatch alone.
Since the predicted spectrum depends on model training configuration, we provide a label-free model selection rule for selecting the near-the-best operation point.
}

\maketitle

\section{Introduction}
\label{sec:introduction}

Machine learning (ML) methods, particularly neural networks, have become a powerful tool for data analysis in physics.
Their strength lies in the ability to identify complex patterns in the data that may be overlooked by, or are difficult to implement within, traditional rule-based algorithmic frameworks.
Rather than relying on a handful of hand-crafted variables, a neural network learns the mapping from the raw detector output to the quantity of interest directly from the data, exploiting correlations between many low-level observables at once.
As a result, ML substantially improves the accuracy of data analyses, prompting its adoption across different applications.

However, the same sensitivity to subtle features of the data that drives this success is also the primary limitation of neural networks.
Namely, a model may latch onto patterns that are specific to the training data set rather than onto physically meaningful structures~\cite{novak2018sensitivity, zhang2024feature}.
Effectively, a network learns to interpolate within the feature phase space defined by its training data.
Applying it to data that lie outside this region constitutes an extrapolation, which is notoriously unreliable.
Such over-specialization introduces biases that may be large and are, by construction, uncontrolled: the network's output carries no indication that the input lies outside the region in which the model was trained.
Moreover, the sensitivity that makes a network accurate also makes it responsive to ever smaller differences between the training and application data, so that the more powerful the model, the more fragile its predictions under such a mismatch.

This problem is of particular importance in physics, where models are typically trained on Monte Carlo (MC) simulations and subsequently applied to experimental data~\cite{farahani2021brief, kouw2018introduction, xu2022eyes, bousmalis2018using, liu2021towards, kushibar2021transductive, wang2022embracing}.
MC simulations unavoidably differ from real-world data, as they cannot perfectly replicate every physical process, detector response, and noise source.
These discrepancies are further compounded by the systematic uncertainties of the simulation itself, which stem from an imperfect knowledge of the underlying physics.
For example, in astrophysics the muon puzzle --- a long-standing discrepancy between the muon content of simulated and observed air showers --- remains unresolved~\cite{ArteagaVelazquez:2023fda, abbasi2022density, aab2015muons, aab2016testing, TelescopeArray:2018eph}.
Such mismatches lead to systematic shifts between the training (MC simulations) and target (experimental data) domains, potentially compromising the physical interpretation of the results.

Domain adaptation (DA) is a broad class of techniques designed to make ML models
robust against systematic shifts between domains~\cite{farahani2021brief,
kouw2018introduction, long2015learning, glorot2011domain, Ganin:2016,
ajakan2014domain}.
In its standard formulation one is given labelled data from a \emph{source}
domain, distributed as $p_\mathrm{src}(x, y)$, and unlabelled data from a
\emph{target} domain, $p_\mathrm{tgt}(x, y)$, and the goal is to learn a
predictor that is accurate on the target.
The most widely used family of methods is adversarial: an auxiliary classifier,
the \emph{adversary}, is trained to tell the two domains apart from the internal
representation $z$ learned by the network, while the network is trained to make
this impossible~\cite{Ganin:2016}.
At convergence the representation is expected to retain the information required
for the task while discarding the information that identifies the domain.
Such methods have been successfully applied in high-energy
physics~\cite{Ghosh:2021roe, Stein:2022nvf, Estrade:2019gzk, Dellaiera:2023txi},
including at the LHC~\cite{Baalouch:2019fhm, CMS:2024zqs, May:2022lhr}, and in
cosmology~\cite{Moss:2018tug, Swierc:2024gdu}.

Domain adaptation rests on two assumptions.
The first is that the domains differ only by a nuisance: the mapping from the
data to the quantity of interest is shared,
$p_\mathrm{src}(y|x) = p_\mathrm{tgt}(y|x)$, and what changes is confined to
features that carry no information about $y$.
The canonical illustration is the MNIST-M benchmark~\cite{Ganin:2016}, in which
the label is the digit and the domain shift is the coloured background.
Since
the background is irrelevant to the digit, a background-invariant representation
transfers across the domains at no cost in accuracy.
The second assumption concerns the mechanism of the alignment.
The adversary sees only the \emph{marginal} distribution of the representation,
which also encodes the distribution of the label.
Enforcing $p_\mathrm{src}(z) = p_\mathrm{tgt}(z)$ is therefore label-preserving
only if the quantity of interest is distributed identically in the two domains,
$p_\mathrm{src}(y) = p_\mathrm{tgt}(y)$~\cite{tachet2020domain}.
Otherwise the alignment can be achieved only by distorting the predictions
themselves.

In physics, neither of these assumptions typically holds.
First, simulations can be wrong about the physics itself, hence the mismatch is not a nuisance but a genuine difference in the relation between the data and the quantity of interest.
Second, in many physics applications the distribution of the target quantity is not shared between the domains, and is in fact the very thing being measured. 
For example, in cosmic-ray energy reconstruction it is the experimental energy spectrum, and in cosmology the redshift distribution.
Since the adversary cannot distinguish a nuisance difference from a difference in the distribution of the label, applying DA naively in such a case may spuriously align the experimental spectrum with the simulated one, biasing the measurement that DA was introduced to protect.

In this paper we perform a systematic study of domain adaptation effects on a toy simulated experiment mimicking an air-shower detector array.
The benchmark exposes three distinct types of mismatch: a misspecification of the simulated physics (the lateral distribution function, LDF), an artifact of the simulated detector response (the waveform shape), and a spectrum (label) shift.
We find that DA performs well as long as the two domains share the same energy spectrum.
However, once spectra are different (label shift), the standard domain-adversarial neural network (DANN)~\cite{Ganin:2016} drags the reconstructed spectrum towards the simulation prior.
This introduces bias in the very quantity the analysis sets out to measure.

To resolve this, we introduce \emph{Adaptive Domain Adaptation} (ADA), which targets the label shift directly.
ADA dynamically assigns weights to the simulated events entering the adversarial loss, so that the source (MC) and target (experimental data) spectra, as estimated by the network itself, coincide.
The spectrum is thereby removed from the set of domain-discriminative features, and the adversary is left to act only on the genuine physical difference between the domains.
Similar importance-weighted adversarial alignment has previously been applied to classification tasks to account for class imbalance~\cite{tachet2020domain, zhang2018importance}.
Our ADA extends this approach to the regression case, where the weights have to be inferred on the fly, in the absence of known target labels.

Mismatches between simulation and experimental data frequently manifest across different stages of the signal processing pipeline. 
We demonstrate that when a domain shift is localized, domain adaptation employing a local domain classifier effectively strips away the detector-level nuisance. 
However, when discrepancies are non-localized or affect global event properties, such approach does not resolve the domain shift and one should use standard, global domain adaptation.

To maximize performance when the true target spectrum is unknown, we also propose \emph{iterative domain adaptation}. 
By feeding the target spectrum predicted in one training round back into the next as an improved prior, the framework iteratively updates its spectrum assumptions. 
This self-consistent refinement mitigates reliance on poorly specified initial priors and leads to better energy spectrum reconstruction.

Our second result addresses the question of how to select a model when no target labels exist.
Adversarial training has a free parameter, the adaptation strength, and the quality of the result may depend on it sharply. 
Yet no quantity available at training time indicates whether the adaptation has succeeded, or is too weak or too strong.
We propose a label-free selection rule based on the cross-seed spread of the network's predictions: model configurations whose independently seeded runs agree with one another are those in which adversarial training has properly converged.
On our toy experiment this rule returns, in every case we studied, a configuration within $25\%$ of the label-chosen optimum. This makes domain adaptation usable in practice in the most challenging case when no reliable target labels exist.

The rest of the paper is structured as follows.
Section~\ref{sec:setup} introduces the toy simulated experiment, the network architecture, and the diagnostic quantities used throughout the paper.
In section~\ref{sec:pure} we consider each of the domain mismatches one at a time, to understand the role of each shift in isolation.
Section~\ref{sec:compound} switches them on together and identifies the failure modes of standard DA.
Adaptive domain adaptation, ADA, is introduced in section~\ref{sec:combined} on the example of a joint LDF and spectrum mismatch.
Section~\ref{sec:wfspec} considers the waveform-plus-spectrum mismatch, and section~\ref{sec:triple} covers the realistic case in which all three shifts are present at once.
We formulate and verify the general label-free model selection rule in section~\ref{sec:stop}.
Finally, section~\ref{sec:summ} concludes the paper.

\section{Data and neural network}
\label{sec:setup}

\subsection{Simulated experiment and datasets}
\label{sec:event}

Our air-shower detector is a $3\times 3$ grid of nine detector stations, with grid spacing $D = 1500$~m.
Each \textit{event} is a single shower observation whose core is placed at a random core-to-center offset drawn uniformly $r_\mathrm{core} \in [200, D/3]$~m.
The azimuth and zenith angles of the air shower are uniformly distributed in $[0, 2\pi)$ and $\cos\theta \in [\cos \pi /4, 1]$, respectively.
The zenith angle is a physical parameter of the event: it fixes the perpendicular distances from the stations to the shower axis, and hence the signal each station registers.
Each event is labelled by its dimensionless energy $E = \log_{10}(E_{\text{ph}}/1\text{EeV}) \in [0,1]$, and we sample energy spectra from Beta distributions.

\begin{figure}
\centering
  \begin{subfigure}[t]{0.46\textwidth}
  \center{\includegraphics[width=1.\linewidth]{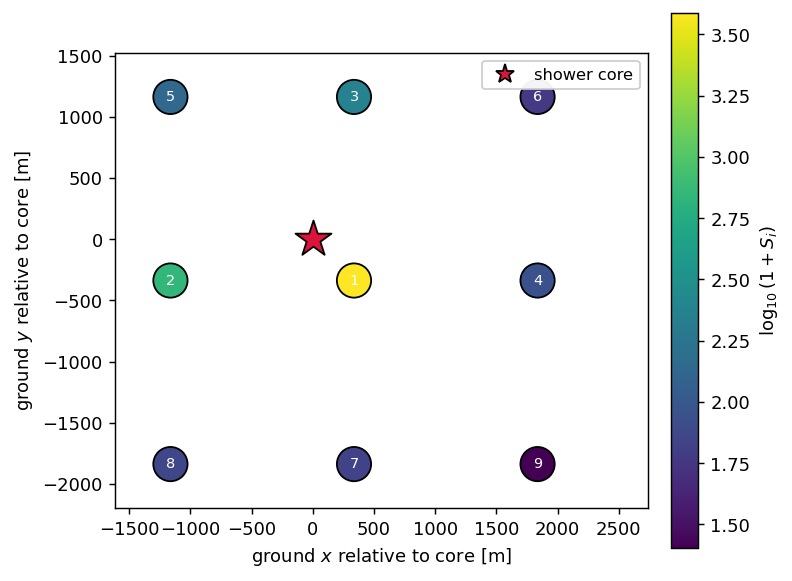}}
  \caption{}
  \end{subfigure}
\hfill
  \begin{subfigure}[t]{0.46\textwidth}
  \center{\includegraphics[width=1.\linewidth]{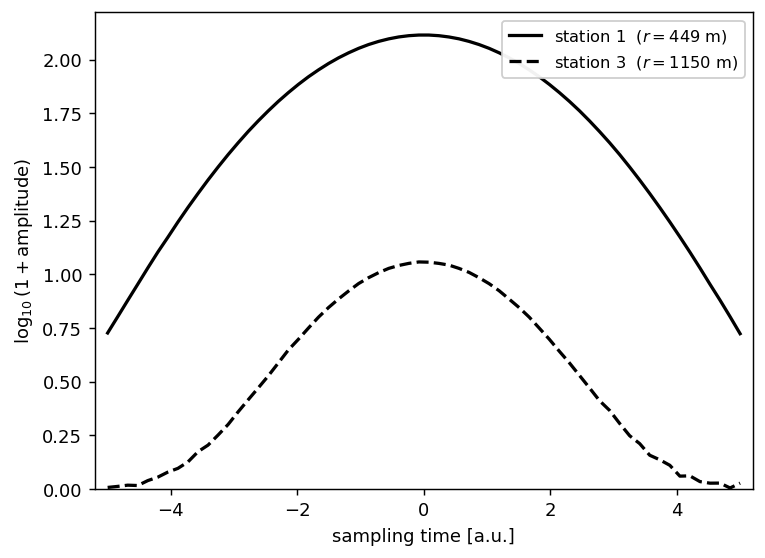}}
  \caption{}
  \end{subfigure}
\caption{An example of an event. (a): station locations and registered signal, (b): example of two waveforms registered by stations, with time axis in arbitrary units (a.u.).}
\label{fig:event_example}
\end{figure}

Each station records a \textit{waveform} --- a time series of the registered signal, sampled in 64 bins that uniformly cover the interval $[-5, 5]$ in arbitrary time units (a.u.).
All nine stations register a signal in every event, and no trigger condition or threshold is applied.
We assume that the waveforms have a Gaussian shape centered at their middle, with the integrated registered signal $S(r, E)$ following a power-law lateral distribution function (LDF):
\begin{equation}
\label{eq:ldf}
    S(r, E) \;=\; S_0(E) \,\Big(\frac{r}{r_0}\Big)^{-\beta_\mathrm{LDF}}\,,~~~~
    S_0(E) \;=\; 100\cdot 10^{E}\,,~~~~ 
    r_0 = 800~\mathrm{m}\,, ~~~~
    \beta_\mathrm{LDF} = 3.0 \,,
\end{equation}
where $r$ is the perpendicular distance from the station to the shower axis and $\beta_\mathrm{LDF}$ is the LDF slope parameter.
The waveform width depends on the signal magnitude through $\sigma(S) = 0.8 + 0.4\,\log_{10}(1 + S)$~a.u., and independent Gaussian noise of standard deviation $0.02$ is added to each bin.
The physical information about the energy thus reaches the network through the integral of the waveform, while its shape and width are set by the simulation of the detector response.
An example of an event is shown in figure~\ref{fig:event_example}.

Besides the waveform samples, each station is described by four scalars: the perpendicular distance to the shower axis $r/r_0$, the ground coordinates $(x, y)/r_0$ with respect to shower core, and the normalized plane-front arrival time $t$.
In the experiments without a waveform mismatch we supply one more feature, the integral of the registered waveform, the \textit{integral bypass}.
All waveform bins and the integral charge are transformed as $\log_{10}(1+S)$, since they span several orders of magnitude in raw values.

To quantify success or failure of DA, on the target domain we evaluate the mean absolute error (MAE) of the energy predictions and the Wasserstein-1 distance (WD) between the predicted, $\hat p_\mathrm{tgt}(E)$, and true target spectra, $p_\mathrm{tgt}^\mathrm{true}$.
WD, also known as the earth mover's distance, is the minimal cost of transforming one distribution into the other, the cost of moving a unit of probability mass being the distance over which it is moved.
The two metrics are complementary: MAE measures the accuracy of individual events, while WD measures the accuracy of the reconstructed spectrum as a whole.
All reported metrics are averaged over five independent trainings that differ only in the initial neural network state (random seed).

In our experiments we will use three datasets, each playing a different role.

The first dataset plays the role of the MC-simulated events that can be used to train the neural network.
We will call it the \textit{training} set $\mathcal{D}_\mathrm{tr}$, with $n_\mathrm{tr} = 40\,000$ labelled events of the \textit{source} (MC) domain.
Their energy distribution is chosen to be flat, $ E_{\text{tr}} \in [0, 1]$, to reduce the energy-dependent bias of event reconstruction and to cover the whole energy range within a single dataset.

The second dataset is the \textit{target}, or ``experimental'' dataset $\mathcal{D}_\mathrm{tgt}$ with $n_\mathrm{tgt} = 30\,000$ unlabelled events, containing the events of physical interest.
In the general case, its spectrum is unknown or known only approximately, and it can differ from the source dataset in event characteristics.
For example, imperfect MC simulations and unknown high-energy physics parameters may result in a different LDF slope, while a mis-simulation of the detector response may result in different waveform shapes.

The third dataset is the \textit{DA-source} dataset $\mathcal{D}_\mathrm{da}$ with $n_\mathrm{da} = n_\mathrm{tgt} = 30\,000$, which is used for domain adaptation alongside the target dataset.
It is generated with the same source-side parameters as $\mathcal{D}_\mathrm{tr}$, but its spectrum is drawn from the approximated target spectrum.
This approximates the condition under which adversarial alignment is safe~\cite{tachet2020domain}, letting the network train over the full energy range while adaptation acts on the genuine domain shift alone.
The spectrum of $\mathcal{D}_\mathrm{da}$ thus plays the role of the experimenter's prior on the target spectrum.

\subsection{Neural network and training}
\label{sec:arch}

Reconstructing the shower energy requires processing the data at two levels: the waveform recorded by each station, and the event as a whole, formed by all nine stations together.
Our network is organized accordingly.
A waveform encoder first summarises each station's waveform into a compact feature vector, and a transformer then combines the per-station information into a single event-level representation, from which the energy is estimated.
Below we describe these two components.

The waveform encoder is a small one-dimensional convolutional neural network (CNN) that maps the 64-bin waveform to a $6$-dimensional feature vector through the sequence
\begin{equation}
\label{eq:wfenc}
(1 \times 64)
\;\xrightarrow{\text{conv}}\; (8 \times 32)
\;\xrightarrow{\text{conv}}\; (16 \times 16)
\;\xrightarrow{\text{conv}}\; (16 \times 8)
\;\xrightarrow{\text{flatten}}\; \mathbb{R}^{128}
\;\xrightarrow{\text{linear}}\; \mathbb{R}^{6} ,
\end{equation}
where $(c \times \ell)$ denotes $c$ channels of length $\ell$.
Each convolution has kernel size $5$ and stride $2$, and every layer is followed by a GELU nonlinearity.
The resulting feature vector encodes the pulse information, and the encoder is trained jointly with the rest of the network to learn the features most useful for energy reconstruction.

The waveform feature vector is then concatenated with the scalar station characteristics: the station coordinates, the plane-front arrival time, and, when the integral bypass is enabled, the integral charge.
A learnable linear layer maps each station vector to the model dimension $d = 32$, producing the sequence of station representations that the rest of the network operates on.

Transformers~\cite{Vaswani:2017} are a natural choice for the event-level part of the model.
Their self-attention mechanism takes into account correlations between all stations, thus potentially enabling more accurate analysis than convolution or recurrent neural networks.
Each event is represented as an unordered set of stations, forming the input to the neural network.
We use a two-layer transformer encoder with two attention heads and a learnable classification token~\cite{devlin2019bert}, which we call the \textit{event summary}, aggregating the station representations into a single vector.
The event summary is read by a two-layer fully connected regression head that outputs the estimated energy, trained with the mean-square error on $E$.
In total the network has $23\,500$ trainable parameters.

\begin{figure}
\centering
\includegraphics[width=0.92\linewidth]{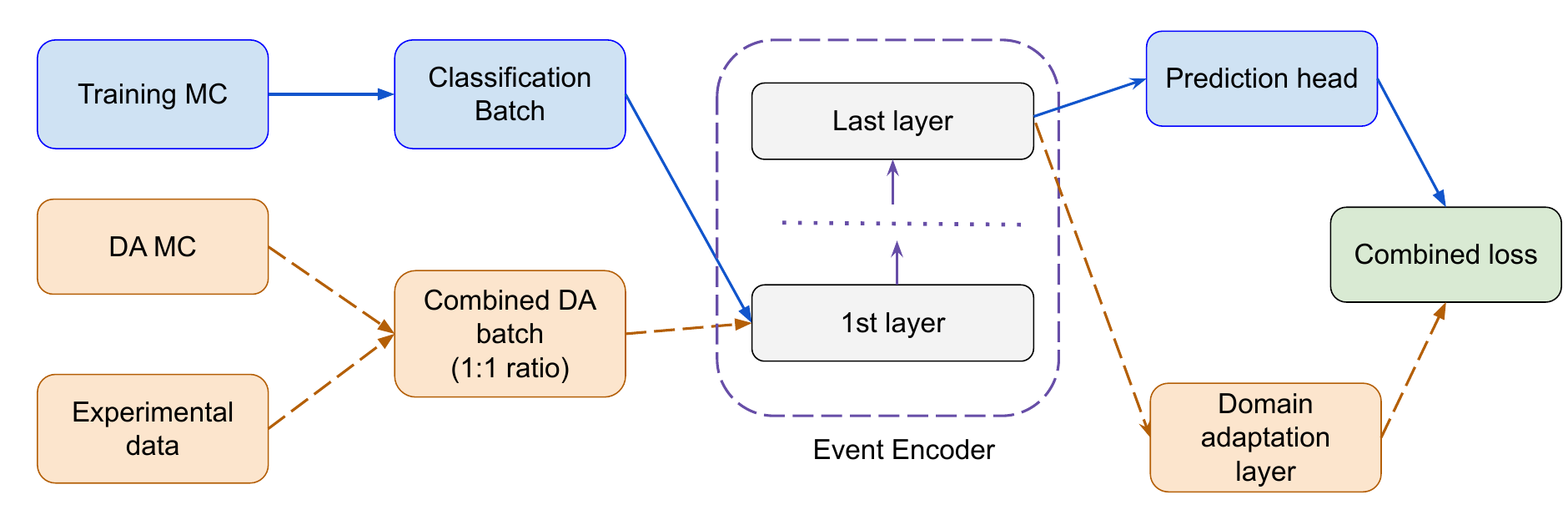}
\caption{Data flow during training and the network architecture.
The labelled training set $\mathcal{D}_\mathrm{tr}$ forms the regression batch (blue), which is processed by the transformer encoder and read by the regression head.
The DA-source $\mathcal{D}_\mathrm{da}$ and the target $\mathcal{D}_\mathrm{tgt}$ are combined in equal proportion into the domain-adaptation batch (orange), which is processed by the same encoder and read by the domain-adaptation layer.
The two paths meet in the combined loss.}
\label{fig:dataflow}
\end{figure}

To make the predictions robust against the domain shift, we attach a \textit{domain-adaptation layer} to the event summary, figure~\ref{fig:dataflow}, following the DANN architecture~\cite{Ganin:2016}.
It comprises a domain classifier, or \textit{critic}, trained to distinguish source and target domains, and a gradient-reversal layer (GRL) inserted between the classifier and the event encoder.
The event-wise critic reads the $32$-dimensional event summary and consists of a linear layer of width $32$, followed by layer normalization, a GELU nonlinearity, and a linear layer producing a single logit.
It searches for the features that best distinguish the two domains.
During backpropagation, the GRL reverses the sign of the critic's gradient, scaled by a factor $\lambda$, before it reaches the encoder.
The encoder is thereby driven to remove precisely the features the critic relies on.
The intended outcome is that the domain can no longer be inferred from the event summary, so that the energy is read from features shared by the two domains.

The domain shift may also be sought at the level of individual stations rather than the whole event.
For this purpose we use a second, waveform-wise critic, attached through its own GRL to the $6$-dimensional output of the waveform encoder.
It is a linear layer of width $32$, followed by a GELU nonlinearity and a linear layer producing a single logit, and it is applied to each station separately.
Both critics are trained with the binary cross-entropy loss.
Unless stated otherwise, domain adaptation is performed with the event-wise critic alone.

The three datasets of section~\ref{sec:event} enter the network as two batches per training step, as shown in figure~\ref{fig:dataflow}.
A \textit{regression batch} is drawn from the labelled training set $\mathcal{D}_\mathrm{tr}$; it passes through the encoder and the regression head and contributes the energy loss.
A \textit{domain-adaptation batch} combines the DA-source $\mathcal{D}_\mathrm{da}$ and the target $\mathcal{D}_\mathrm{tgt}$ in equal proportion; it passes through the encoder and the domain-adaptation layer and contributes the domain loss.
By summing these two losses, the network jointly optimizes for accurate energy mapping via the training set and domain invariance via the unlabelled datasets.

Optimizing this joint loss requires carefully managing the adversarial training dynamics. 
The shared encoder must balance opposing updates: the standard gradient minimizing the regression loss and the reversed gradient maximizing the critic's confusion. 
We found that adaptive optimizers like Adam perform poorly under these conditions, as their coordinate-wise scaling disproportionately amplifies the conflict between the two gradients. 
Compared to our preferred optimizer, Adam degrades target recovery by a factor of three and lowers run-to-run robustness. 
Although plain stochastic gradient descent (SGD) handles the gradient conflict better, it converges slowly. 
To achieve stable and timely convergence, we utilize SGD with a momentum of 0.9, which successfully damps the oscillations inherent to the adversarial loss landscape.

The network is trained for 60 epochs.
The adaptation strength $\lambda$ is ramped up from zero following the standard DANN schedule~\cite{Ganin:2016},
\begin{equation}
\label{eq:lambda}
\lambda(p) \;=\; \lambda_\mathrm{max}\left(\frac{2}{1+e^{-\gamma p}} - 1\right),
\qquad \gamma = 10,
\end{equation}
where $p \in [0,1]$ is the fraction of the training completed.
This lets the regression head learn a reasonable energy mapping before the
adversarial pressure is applied, which improves the stability of the training.
The asymptotic value $\lambda_\mathrm{max}$ is the only free parameter of the
adaptation.

\section{Single-axis shifts}
\label{sec:pure}

We start by considering each of the domain mismatches separately: the simulation artifact (waveform shape), the physical simulation mismatch (LDF slope), and the energy spectrum (label shift).
The true target spectrum is fixed to be $\mathrm{Beta}(2,5)$, figure~\ref{fig:warmup_beta}, and we vary DA-source spectrum to model the label shift.
Throughout the paper we report metrics of the model chosen by our label-free selection rule, which is formulated in section~\ref{sec:stop}, when all of the ingredients for mitigating complex domain shifts are in place.

\subsection{Warm-up: a waveform-shape mismatch}
\label{sec:warmup}

\begin{figure}[t]
\centering
\begin{subfigure}[b]{0.32\textwidth}
\centering
\includegraphics[width=\linewidth]{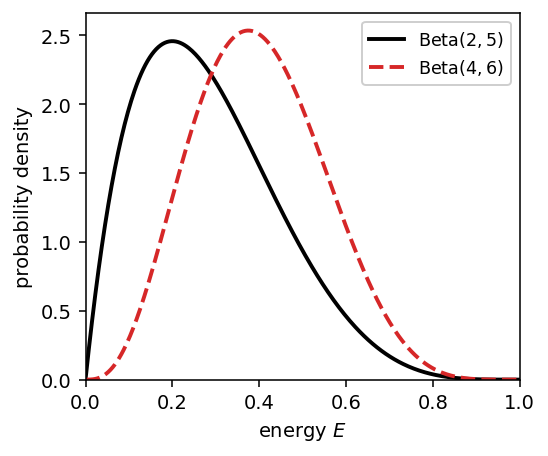}
\caption{}\label{fig:warmup_beta}
\end{subfigure}
\hfill
\begin{subfigure}[b]{0.32\textwidth}
\centering
\includegraphics[width=\linewidth]{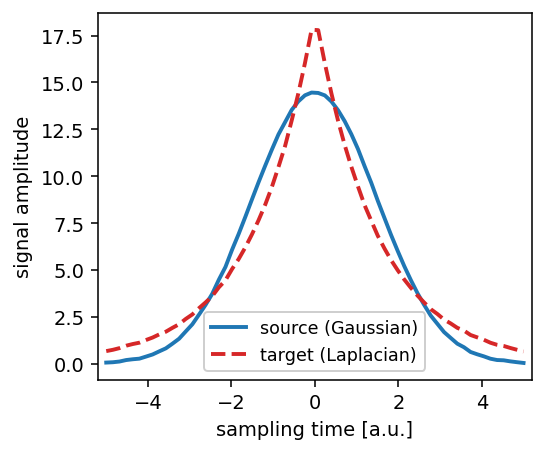}
\caption{}\label{fig:warmup_wf}
\end{subfigure}
\hfill
\begin{subfigure}[b]{0.32\textwidth}
\centering
\includegraphics[width=\linewidth]{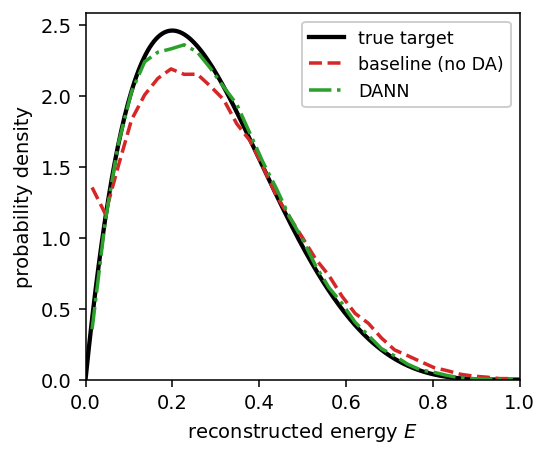}
\caption{}\label{fig:warmup_spec}
\end{subfigure}
\caption{\textbf{(a)} The true target spectrum, $\mathrm{Beta}(2,5)$, and the $\mathrm{Beta}(4,6)$ spectrum used for the DA-source where a label shift is present.
\textbf{(b)} A source (Gaussian) and target (Laplacian) waveform carrying the same integrated charge but a different pulse shape.
\textbf{(c)} The true target spectrum and the target spectrum reconstructed without domain adaptation (baseline) and with DANN.}
\label{fig:warmup}
\end{figure}

We begin with a mismatch in the recorded waveform shape, a simulation artifact of the detector response.
The source and target share the energy spectrum and the lateral distribution function, and differ only in the pulse shape: Gaussian in the source and Laplacian in the target, figure~\ref{fig:warmup_wf}.
In both cases, the integral of the pulse is the same, making the shape a nuisance parameter.
The network has to learn to read out the waveform integral in a domain-invariant way, which is a classical DANN application, allowing us to demonstrate how it works on a simple physical example.

A neural network trained on the labelled source $\mathcal{D}_\mathrm{tr}$ alone, to which we will refer as the \textit{no-DA baseline}, reconstructs the source energy accurately: its mean absolute error on the source domain is $\mathrm{MAE}_\mathrm{src} = 0.005$.
Its predictions on the target, however, are biased.
The reconstructed spectrum is shifted away from the truth, $\mathrm{WD}_\mathrm{tgt} = 0.055$, and the per-event error grows tenfold, $\mathrm{MAE}_\mathrm{tgt} = 0.054$.
The reference point for these numbers is the \textit{matched-domain floor} --- the value a metric takes when the target is generated with the same parameters as the source.
It is set by the detector noise and by the resolution of the reconstruction, and no adaptation can bring a metric below it.
Here the floor is $\mathrm{WD} = 0.003, ~ \mathrm{MAE} = 0.0054 $, so the baseline sits almost twenty times above it.

The bias arises because the network reads the energy from features that are specific to the source pulse shape.
It may rely on the peak height, on the pulse width, or on any other combination of the samples that reproduces the integral on the source data, and which of them a particular training settles on is not reproducible from run to run.
What matters is not the identity of these features but the fact that they are domain-specific, so that their reading does not carry over to the target domain.

The pulse shape is exactly the kind of nuisance the domain-adaptation layer is designed to suppress.
The energy is carried by the integrated charge, which the two domains share, while the shape is all that separates them.
The critic can thus tell source from target only through the shape, and the gradient-reversal layer drives the encoder to discard it, leaving a shape-invariant reading of the integrated charge.
With the domain-adaptation loss switched on, $\lambda = 0.05$, the bias vanishes: the target recovery improves to $\mathrm{WD}_\mathrm{tgt} = 0.005$, near the matched-domain floor, and the per-event error to $\mathrm{MAE}_\mathrm{tgt} = 0.008$, figure~\ref{fig:warmup_spec}.
For this simple case DA works over a wide range of $\lambda$, from $0.005$ to $0.05$.

\subsection{Physics mismatch: LDF slope}
\label{sec:cov}

Another source of domain shift is a physics-related mismatch in MC simulations.
For example, unknown high-energy physics may lead to a systematic, energy-dependent mismatch in the LDF --- how the deposited integrated charge depends on the distance from the station to the shower core.
In this section we model this situation by holding the source slope fixed, while letting the target slope grow with energy:
\begin{equation}
\beta_\mathrm{src} = 3.0 \,, ~~~ \beta_\mathrm{tgt}(E) = 3.0 + 0.5\,E \;.
\end{equation}
As a result, high-energy target-domain air-showers concentrate their signal nearer the axis than source-domain air-showers of the same energy.
Both domains draw from the true $\mathrm{Beta}(2, 5)$ spectrum, so no label shift is present.
In DA terminology, we are considering a parametric conditional shift in $P(x \mid E)$, parametrized by the LDF slope parameter.

A network trained on the labelled source learns the energy-to-signal map implied by the source physics, and reads every target event through it.
Because the target LDF is steeper, the high-energy showers --- where the slope gap is widest --- are mis-read the hardest.
The predicted target spectrum is therefore expected to be compressed toward its low-energy bulk.
We measure the compression by the calibration slope on the target domain,
\begin{equation}
\kappa \;=\; \mathrm{cov}(\hat E, E)\,/\,\mathrm{var}(E) \,,
\end{equation}
which equals unity for a faithful reading and tends to zero for a prediction collapsed onto a constant.
The no-DA baseline has $\kappa = 0.49$ and $\mathrm{WD}_\mathrm{tgt} = 0.143$, more than an order of magnitude above the matched-domain floor.

How can domain adaptation mitigate such a domain shift?
To answer this question, recall that each event is registered by nine stations, at different distances $r$ from the axis.
The energy enters every station through the common amplitude $S_0(E)$, and the slope through a common power law in $r$.
The functional dependence of signal on distance across the event therefore over-determines the pair $(E, \beta)$.
A network that learns this dependence can recover the energy in a slope-independent way.

Domain adaptation is what drives the encoder onto that reading.
By its very definition, the reversed-gradient loss penalises any feature whose distribution differs between source and target.
The source-LDF amplitude pattern is such a feature, and it is the most direct route to a source-only fit, so it is the first to be suppressed.
The slope-tolerant regression draws a smaller penalty from the critic and survives.
The adversarial loss, however, is not the only term acting on the encoder.
The regression loss is applied to the same weights and requires the energy to remain readable, so the training settles at a compromise: a domain-discriminative feature is removed only insofar as the regression can afford to lose it.
Features that separate the domains but are needed to read the energy are preserved, and we return to the consequences of this below.

\begin{figure}[t]
\centering
\begin{subfigure}[b]{0.45\textwidth}
\centering
\includegraphics[width=\linewidth]{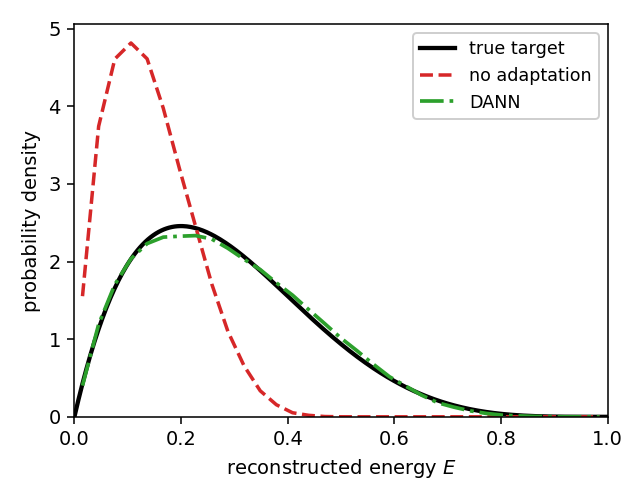}
\caption{}\label{fig:tc1_spectra}
\end{subfigure}
\hfill
\begin{subfigure}[b]{0.45\textwidth}
\centering
\includegraphics[width=\linewidth]{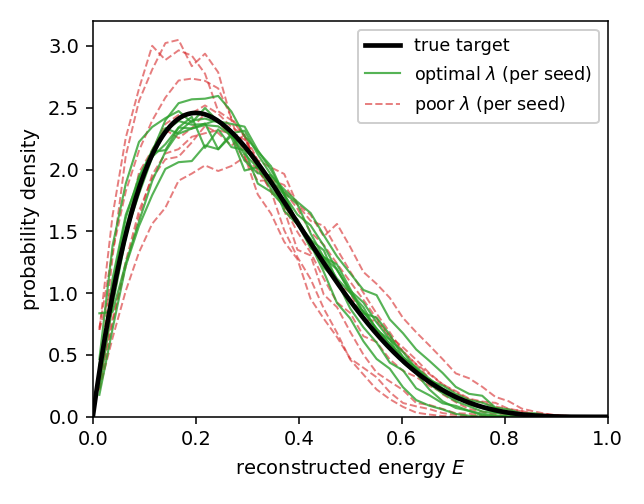}
\caption{}\label{fig:tc1_seed}
\end{subfigure}
\caption{Pure lateral-distribution mismatch.
\textbf{(a)} The predicted target spectrum with no adaptation and with DANN at the adopted recipe, against the true $\mathrm{Beta}(2,5)$ target.
\textbf{(b)} Per-seed predicted target spectra at the optimal $\lambda$ and at a poorly-chosen (too-large) $\lambda$.}
\label{fig:tc1}
\end{figure}

The parameter $\lambda$ of the gradient-reversal layer controls how strongly the adversary pushes relative to the regression loss.
Too weak a $\lambda$ leaves the source-LDF reading in place, and the target energy predictions stay biased.
Across a broad window, $\lambda \approx 0.04$--$0.09$, the recovery is near-complete and stable.
At the optimal point it reaches $\kappa = 0.97$ and $\mathrm{WD}_\mathrm{tgt} = 0.014$, a factor-of-ten improvement over the no-DA baseline, figure~\ref{fig:tc1_spectra}.
At larger $\lambda$ the predictions start to diverge from the true target spectrum, and the spread between trainings with different random seeds grows, figure~\ref{fig:tc1_seed}.
These training-to-training fluctuations are the reason why the reported metrics are averaged over several seeds.

We observed that the critic's own loss does not flag whether domain adaptation was successful or not.
In both cases, the loss saturates at $\ln 2$, the value corresponding to indistinguishable domains.
Thus, optimal adaptation strength cannot be chosen from the adversarial loss alone.

The considered case reveals that successful DA does not imply that the source and target representations become indistinguishable by the end of the training.
Although the critic saturates at a loss of $\ln 2$, a logistic regression freshly fitted to the frozen event-summary vectors does learn to separate the domains.
We fit it on one half of the source and target events and quote its accuracy on the other half.
For the no-DA baseline it reaches $\mathrm{LR}_\mathrm{acc} = 83\%$, while for the optimal DA configuration the accuracy \emph{rises} to $94\%$.
This is not a failure of the training, but the compromise between energy reconstruction and domain adaptation goals at work.
To recover the energy slope-tolerantly, the encoder must read the local shape of the LDF, so the latents necessarily carry the slope, and with it the domain.
A literally domain-invariant representation is neither what the physics permits nor what it needs.
What adaptation makes invariant is the \emph{prediction} --- the encoder and the regression head learn to use the slope to infer the right energy in both domains.

\subsection{Label shift: an energy-spectrum mismatch}
\label{sec:spec}

We now consider a pure \emph{label shift}: the source and target share the detector response $P(x \mid E)$, and only their energy spectra differ.
This corresponds to the situation in which simulations perfectly reproduce real experimental conditions, while the quantity of interest --- the energy spectrum in our case --- is unknown or known only very approximately.
We draw the DA-source dataset from $\mathrm{Beta}(4,6)$, while the target keeps the true $\mathrm{Beta}(2,5)$ spectrum and the labelled training set stays flat.

Since there is no physical mismatch, the no-DA baseline learns the correct mapping from detector responses to energy labels on the training data.
Hence it reconstructs a target event as accurately as a DA-source event: the physics is identical, and the network only interpolates.
Without adaptation it reaches $\mathrm{WD}_\mathrm{tgt} = 0.003$, at the matched-domain floor.

Enabling domain adaptation degrades the predictions.
The only quantity that separates the two domains is the energy spectrum, so the critic reads the spectrum as the domain-discriminating feature, and the gradient-reversal layer drives the encoder to make the encoded energy indistinguishable between the domains.
This pull is opposed by the regression loss, which fixes the correct energy mapping over the whole energy range of the training dataset.
At the equilibrium point the predictions end up slightly shifted towards the DA-source spectrum, which introduces a bias.
The distortion grows with $\lambda$: at $\lambda = 0.05$ the target distance is four times the floor, $\mathrm{WD}_\mathrm{tgt} = 0.009$.
For $\lambda \geq 0.15$ the training collapses --- the adversarial pressure overwhelms the regression head, the source error grows by an order of magnitude, and the predicted energies lose any relation to the true ones.
Under a pure label shift, therefore, the correct action is not to adapt at all.

This result naturally raises the question of how to identify if domain adaptation is needed.
The difficulty is that the energy spectrum is a domain-discriminative feature by itself, so a difference in any diagnostic quantity --- the similarity of the event summaries across the domains, for instance --- may be caused by the spectrum alone.
Moreover, since the network is trained on the simulated source data, it may map the target events onto the same manifold the source events live on, thus hiding a difference that is in fact present.

To avoid these traps, we use a test following the logic of foundation models~\cite{devlin2019bert}.
We train an autoencoder on the source domain only: the event encoder of section~\ref{sec:arch} is followed by a decoder that reconstructs the full event --- all per-station characteristics, including waveform encodings --- from the event summary, with the mean-square error as the reconstruction loss.
The trained autoencoder is then applied to the target domain, and we compare the reconstruction error there with the error on held-out source events.
If the ratio of the two is close to unity, the target events lie on the same manifold as the source events, and the domain shift, if present, is minimal.
A genuine shift of the detector response or of the physics pushes the target off that manifold, and the error rises sharply.
Across the three single-axis cases the target-to-source ratio is $1.0$ for the spectrum shift of this section, and approximately $20$ for the LDF and waveform shifts of sections~\ref{sec:cov} and~\ref{sec:warmup}, respectively. 
A simple threshold therefore separates the mismatches that call for adaptation from the label shift, which does not. 	

\section{Compound shifts}
\label{sec:compound}

A discrepancy between simulation and data is rarely confined to one axis.
We now switch the shifts on together --- first in pairs, then all three at once.
When the two mismatches are both conditional (LDF and waveform shifts), section~\ref{sec:waveform}, standard domain adaptation recovers the target spectrum.
However, once there is an energy spectrum mismatch, standard domain adaptation fails.
Sections~\ref{sec:combined} and~\ref{sec:wfspec} consider these cases and introduce the solution --- adaptive domain adaptation.
In section~\ref{sec:triple} we consider all three mismatches together.

\subsection{Waveform and LDF slope mismatch}
\label{sec:waveform}

We start with the two conditional shifts acting together: a waveform shift, with Gaussian pulses in the source and Laplacian in the target, and an LDF slope mismatch, with $\beta_\mathrm{tgt}(E) = 3.0 + 0.5\,E$.
Both energy spectra are held at $\mathrm{Beta}(2,5)$.

The no-DA baseline compresses the target energy predictions, with a calibration slope of $\kappa = 0.53$ and a per-event error of $\mathrm{MAE}_\mathrm{tgt} = 0.15$.
Both mismatches contribute to this: the steeper target LDF makes the network under-read the high-energy showers, while the sharper target pulse carries the same integrated charge in a different distribution of the samples.

A single critic attached to the event summary mitigates the domain shift.
Across a broad window of the adaptation strength, $\lambda_\mathrm{ev} \approx 0.005$--$0.02$, the recovered spectrum spans the true energy range and varies little between random seeds.
At the optimal configuration, $\lambda_\mathrm{ev} = 0.01$, it reaches $\kappa = 0.84$, $\mathrm{MAE}_\mathrm{tgt} = 0.065$ and $\mathrm{WD}_\mathrm{tgt} = 0.009$.
The remaining bias is most pronounced for highest-energy showers, where both mismatches are strongest.

One might expect that adding a per-station waveform critic would improve on this, since the pulse-shape mismatch originates at exactly that level.
This is not the case.
The two critics come with two independent reversal strengths, $\lambda_\mathrm{ev}$ and $\lambda_\mathrm{st}$.
On this two-dimensional grid the best configuration has $\lambda_\mathrm{st} = 0.0$, indicating that two critics might be in conflict, or that a single event critic is enough.

To identify which of the possibilities is the correct one, we tested two different critic architectures: an entangled configuration and a disentangled scheme.
In the default entangled configuration, both the event critic and the waveform critic gradients are allowed to propagate back to the waveform encoder.
In the contrasting disentangled scheme, we blocked the gradient of the event critic to the waveform encoder, forcing the per-station representation to be shaped by the waveform critic alone.
Testing the disentangled scheme resulted in a degradation of $\text{WD}_\text{tgt}$ by a factor of three (figure 4a).
To understand why, note that the station critic reads only the output of the waveform encoder, so it is blind to the distance $r$.
This suffices for a genuine nuisance, as in section~\ref{sec:warmup}, where the two pulse shapes carried the same integrated charge and suppressing the shape left the energy untouched.
An LDF mismatch is not of that kind: at fixed energy the target amplitude differs from the source one by a factor $(r/r_0)^{-\Delta\beta}$, so the same recorded amplitude corresponds to different energies in the two domains.
The per-station difference is thus not a nuisance but an $r$-dependent physics mismatch.

The event summary, in contrast, holds the amplitudes of all nine stations together with their distances, so the slope-tolerant reading of section~\ref{sec:cov} remains available to a critic acting on it.
This is the main result of this section: when several conditional shifts act at once, a single critic that sees all of them is preferable to several specialized ones.
A critic confined to one level cannot tell a nuisance from a genuine difference, and thus can be fooled into false domain-invariance.

\begin{figure}[t]
\centering
\begin{subfigure}[b]{0.32\textwidth}
\centering
\includegraphics[width=\linewidth]{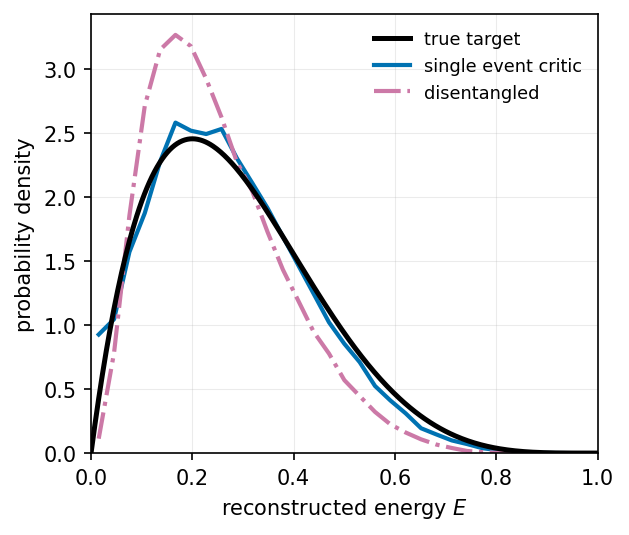}
\caption{}\label{fig:spectra_a}
\end{subfigure}
\hfill
\begin{subfigure}[b]{0.32\textwidth}
\centering
\includegraphics[width=\linewidth]{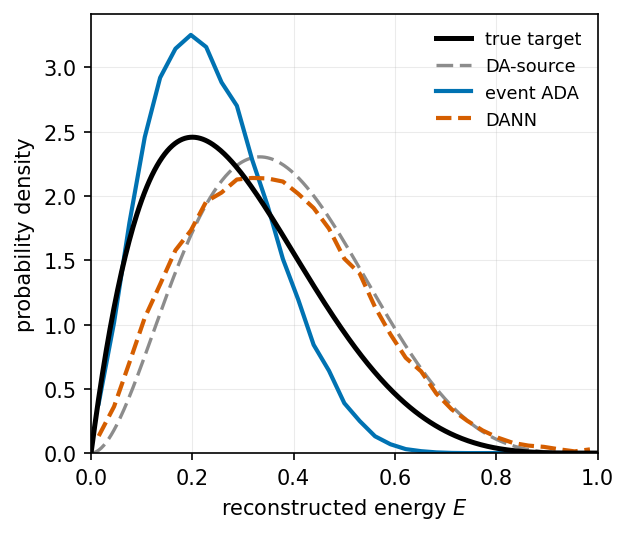}
\caption{}\label{fig:spectra_b}
\end{subfigure}
\hfill
\begin{subfigure}[b]{0.32\textwidth}
\centering
\includegraphics[width=\linewidth]{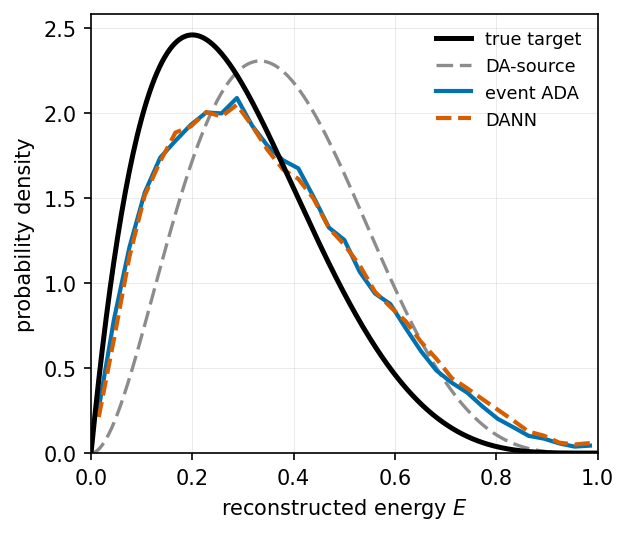}
\caption{}\label{fig:spectra_c}
\end{subfigure}
\caption{Reconstructed target spectra against the true $\mathrm{Beta}(2,5)$ target.
\textbf{(a)} Waveform and LDF mismatch with matched spectra, DA-source and target both $\mathrm{Beta}(2,5)$: single event critic and disentangled two-critic scheme.
\textbf{(b, c)} LDF and spectrum mismatch, true target $\mathrm{Beta}(2,5)$ and DA-source $\mathrm{Beta}(3,5)$, at LDF slope $+0.5E$ (b) and $-0.2E$ (c), comparing ADA and DANN predictions.}
\label{fig:spectra_examples}
\end{figure}

\subsection{LDF slope and spectrum mismatch}
\label{sec:combined}

We proceed with combined LDF slope and energy spectrum mismatches, that is, a physical and a label shift acting together.
The DA-source LDF slope is fixed to $\beta_\mathrm{src} = 3.0$, while the target slope changes with energy as $\beta_\mathrm{tgt}(E) = 3.0 + \alpha E$.
The target keeps the true $\mathrm{Beta}(2,5)$ spectrum, while the DA-source is drawn from $\mathrm{Beta}(4,6)$.
Below we first examine the default configuration, $\alpha = 0.5$, and then sweep both the slope and the DA-source spectrum to map out the behaviour of the methods.

A neural network trained without DA pulls the target energy spectrum to the low-energy tail.
The Wasserstein distance between the true and the predicted spectrum is $\mathrm{WD}_\mathrm{tgt} = 0.143$, with a calibration slope of $\kappa = 0.49$ and a predicted target mean of $0.14$ against the true $0.286$.
This is more than an order of magnitude above the matched-domain floor of $\mathrm{WD} = 0.003$.

A standard DANN fails in this regime.
Across all $\lambda$ values it over-corrects on average, at best reaching $\mathrm{WD}_\mathrm{tgt} = 0.14$ and $\kappa = 1.22$.
The distance to the true spectrum is thus no better than that of the no-DA baseline, while the failure mode is inverted: the baseline compresses the predicted spectrum, and DANN stretches it.

The underlying reason why DANN fails is that the LDF slope and the energy spectrum mismatches are entangled, with no clear way to separate them.
Specifically, the critic's gradient on the encoder is a sum of two pulls.
The first drives the encoder toward LDF-tolerant features, which is useful.
The other drives it toward an energy-distribution-invariant representation, which is harmful, since it explicitly aligns the energy spectra of the two domains.
What is different from the pure label shift of section~\ref{sec:spec} is that the second pull is no longer resisted.
There the two domains were physically identical, so distorting the target predictions meant distorting the same mapping that the regression loss was fixing on the labelled training set, and the regression head held the predictions in place.
Here the target events genuinely look different, and the encoder can move their predictions without influencing the source regression loss.
The anchor is gone, the training is ill-posed, and the predictions are unreliable.

To resolve this, one should stop the adversary from seeing the spectrum at all, so that it can act only on the genuine domain difference --- the LDF response.
If the DA-source and the target carried the same energy spectrum, their predicted-energy marginals would already coincide, and the critic could no longer read the spectrum as a domain feature.
Whatever separation it then finds is the physical shift we actually want it to remove.
This is the condition under which adversarial alignment is known to be safe~\cite{tachet2020domain}.
The spectrum difference can be cast out of the DA loss by reweighting the DA-source events so that the two domains have the same effective (weighted) spectrum.
The question then becomes how to find weights such that the reweighted DA-source spectrum is a faithful approximation of the target spectrum.

We approach this problem as follows.
The DA-source spectrum is considered as a prior of the true target energy spectrum, which is updated after each epoch of neural network training.
This correction is done via assigning weights to DA-source events so that effective spectra, estimated by the neural network, of DA-source and target events are the same.
Specifically, after each training epoch we bin the predicted source and target energy spectra into $B$ uniform bins on $[0,1]$ and assign to a DA-source event whose predicted energy falls in bin $b$ the weight
\begin{equation}
\label{eq:ada_weight}
w_b \;=\; \frac{\hat p_\mathrm{tgt}(b)}{\hat p_\mathrm{da}(b)} \;,
\end{equation}
where $\hat p_\mathrm{\cdot}(b)$ is the fraction of events of the corresponding domain falling into bin $b$.
We further clip values to range of $(0.01, 30)$ to avoid training instabilities due to exploding weights.
This removes the energy-spectrum difference from the DA loss, forcing the critic to search for the genuine domain shift.
The weights are updated after each epoch, so the network converges to a fixed point at which the weights stop changing.

We dub this approach \emph{Adaptive Domain Adaptation} (ADA), summarised in algorithm~\ref{alg:ada}. 
It is similar in spirit to importance-weighted DANN for classification tasks~\cite{tachet2020domain, zhang2018importance, saerens2002adjusting, lipton2018detecting}.
The difference is that ADA generalizes the approach to the regression case, when the predictions are themselves affected by the very shift being corrected.
Note that ADA is self-referential: it assigns weights to events based on its own predictions.
We will see in the sweep below that this leaves a visible imprint on the results, without preventing ADA from bounding the error where standard DA does not.

\begin{algorithm}[t]
\caption{Adaptive Domain Adaptation (ADA).}
\label{alg:ada}
\begin{algorithmic}[1]
\Require Labelled source $\mathcal{D}_\mathrm{tr}$, DA-source $\mathcal{D}_\mathrm{da}$ (unlabelled, source physics, arbitrary energy spectrum), target $\mathcal{D}_\mathrm{tgt}$ (unlabelled, target physics)
\Require Number of energy bins $B$, EMA decay $\eta$, epochs $N$
\State Initialise per-bin weights $w_b^{(0)} \gets 1$
\For{$k = 1$ \textbf{to} $N$}
  \State Train the neural network for one epoch on $\mathcal{D}_\mathrm{tr}$ for regression, and on $\mathcal{D}_\mathrm{da}$ with weights $w^{(k-1)}$ and $\mathcal{D}_\mathrm{tgt}$ for domain adaptation.
    \State At epoch end, get normalized histograms of predicted energy $\hat E$ on source and target domains:\newline $\hat p_\mathrm{tgt}\gets\mathrm{hist}(\hat E(\mathcal{D}_\mathrm{tgt}))$,\quad $\hat p_\mathrm{da}\gets\mathrm{hist}(\hat E(\mathcal{D}_\mathrm{da}))$
    \State Update source-event weights: $\tilde w_b\gets\hat p_\mathrm{tgt}(b)/\hat p_\mathrm{da}(b)$
    \State (optional) Use EMA to reduce oscillations: $w_b^{(k)}\gets\eta\,\tilde w_b+(1-\eta)\,w_b^{(k-1)}$
\EndFor
\State \Return trained encoder and regressor, final ADA weights
\end{algorithmic}
\end{algorithm}

To test ADA, we compare it against standard domain adaptation (DANN), varying two parameters.
The first is the LDF slope coefficient $\alpha$, taking the values $\alpha \in \lbrace 0.5\,, 0.2\,, -0.2 \rbrace$.
This covers a strong and a mild domain shift, while the change of sign reverses the direction of the bias.
For positive (negative) $\alpha$ the signal falls faster (slower) with distance, resulting in underestimation (overestimation) of the energy according to the DA-source physics model.
The second parameter is the difference between the mean energies of the DA-source and target spectra, which we vary from $-0.1$ to $0.3$ by changing the parameters of the Beta distribution the DA-source is drawn from.
We use $B = 10$ bins, so that each energy bin has $n\geq10$ events, and an exponential moving average (EMA) of the weights with decay $0.3$, which damps their oscillations from epoch to epoch.
Both values were chosen for the stability of the training, including the stability of the weight updates themselves. 

Figures~\ref{fig:spectra_b} and~\ref{fig:spectra_c} compare ADA and DANN predictions for the fixed DA-source spectrum $\mathrm{Beta}(3,5)$ (mean difference $+0.09$) and different-sign LDF slope mismatches, $\alpha \in \lbrace 0.5, -0.2 \rbrace$. 
For $\alpha=0.5$, DANN pulls the prediction towards the DA-source spectrum and over-reads, while ADA stays close to the target, whereas at $-0.2E$ the two nearly coincide.

\begin{figure}[t]
\centering
\begin{subfigure}[b]{0.32\textwidth}
\centering
\includegraphics[width=\linewidth]{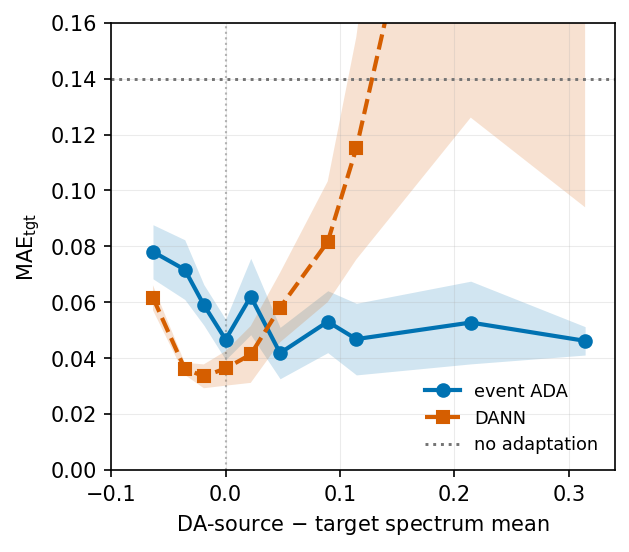}
\caption{}\label{fig:slope_a}
\end{subfigure}
\hfill
\begin{subfigure}[b]{0.32\textwidth}
\centering
\includegraphics[width=\linewidth]{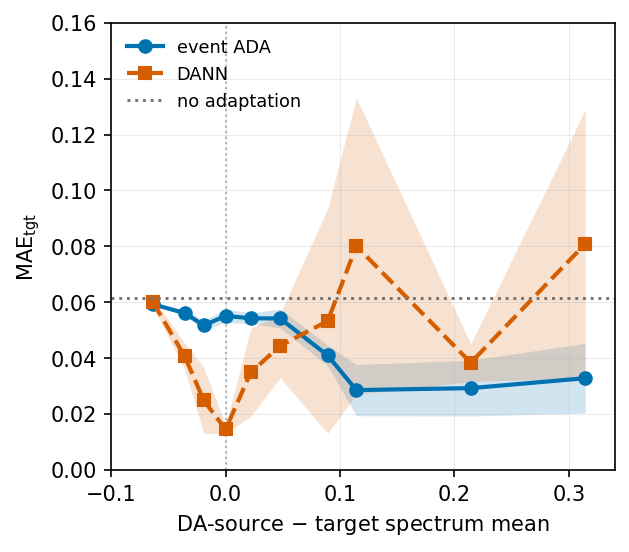}
\caption{}\label{fig:slope_b}
\end{subfigure}
\hfill
\begin{subfigure}[b]{0.32\textwidth}
\centering
\includegraphics[width=\linewidth]{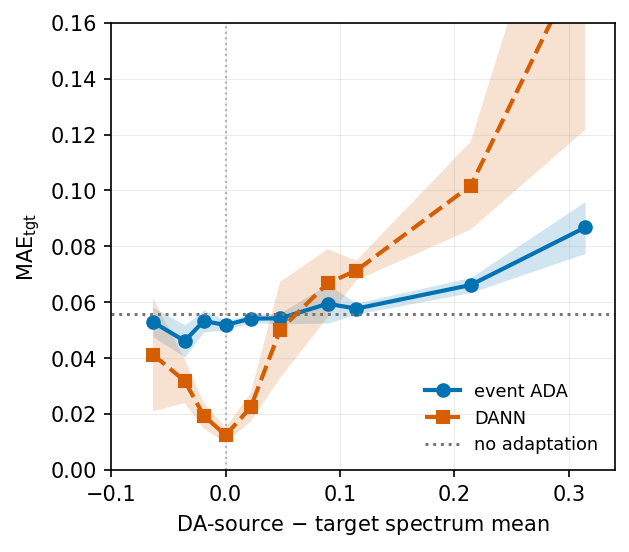}
\caption{}\label{fig:slope_c}
\end{subfigure}
\caption{Target per-event error $\mathrm{MAE}_\mathrm{tgt}$ against the DA-source $-$ target spectrum-mean difference, for event ADA and DANN with the no-adaptation baseline, at LDF slope \textbf{(a)} $+0.5E$, \textbf{(b)} $+0.2E$ and \textbf{(c)} $-0.2E$. Each operating point is chosen by the label-free rule of section~\ref{sec:stop}; bands are $\pm 1$ std over the seeds.}
\label{fig:slope_energy}
\end{figure}

The Wasserstein distance as the function of DA-source and target spectra difference is shown in figure~\ref{fig:slope_energy}.
DANN outperforms ADA in a narrow band where the DA-source spectrum is close to the true target, which lets DANN concentrate on the genuine domain mismatch, while ADA's weights fluctuate, which are further injected into the adversarial loss.
At matched spectra DANN reaches $\mathrm{WD}_\mathrm{tgt} = 0.008$--$0.030$ across the three slopes, against ADA's $\mathrm{WD}_\mathrm{tgt} \approx 0.05$.
At larger spectrum differences ADA performs better --- at a spectrum-mean difference of $0.3$ the DANN error grows to $\mathrm{WD}_\mathrm{tgt} \approx 0.09$--$0.29$, while ADA stays near $\mathrm{WD}_\mathrm{tgt} = 0.03$--$0.09$.
This illustrates that ADA is preferable when the true target spectrum is known only approximately, substantially reducing the prediction bias of standard DA.

Note that ADA behaves differently for positive and negative $\alpha$.
For positive $\alpha$ the error decreases with the spectrum difference, from $0.078$ to $0.046$ at $+0.5E$.
The LDF mismatch makes the network underestimate event energies, so reweighting the DA-source towards higher energies moves the predicted spectrum back towards the truth.
For negative $\alpha$ the situation is reversed.
The network overestimates the energy, which raises the ADA weights for high-energy events, which in turn reinforces the overestimation, and the error grows from $0.05$ to $0.09$ at $-0.2E$.
This is the self-referential estimate of the weights at work, and it is the price of inferring the target spectrum from the predictions of the very network being corrected.
Even in the latter case, however, the loop does not run away: ADA remains stable and yields a reasonable estimate of the true target spectrum.

To summarise, when a physical and a label shift act together, standard domain adaptation is unsafe at every adaptation strength.
Specifically, it biases the predictions in an uncontrolled, mismatch-dependent way.
Removing the spectrum from the adversary's view repairs this.
ADA keeps the error bounded across the whole range of spectrum mismatches we scanned, at the cost of a modest floor in the regime where the assumed spectrum happens to be correct.
In section~\ref{sec:triple} we will see how ADA's predictions can be improved further.

\subsection{Waveform and spectrum shift}
\label{sec:wfspec}

The last pair to switch on combines the waveform-shape shift with a spectrum shift, holding the LDF slope fixed.
The source pulses are Gaussian and the target pulses Laplacian, carrying the same integrated charge at the same energy.
As in section~\ref{sec:combined}, the DA-source and the target draw their energies from different $\mathrm{Beta}$ spectra, and we vary the difference between their means.

The two mismatches enter the problem in different ways.
The waveform shape is the only genuine domain shift: it differs between the domains and must be corrected.
The spectrum difference, on the other hand, does not require correction by itself --- a neural network trained in a domain-invariant way would reconstruct it correctly.
The spectrum becomes a problem only because the adversary sees it and treats it as a domain feature.
Disentangling the two --- letting the adversary act on the genuine mismatch alone --- should therefore make the label harmless, and give ADA a better recovery.
These motivate the two approaches tested below.

The first follows the logic of foundation models~\cite{devlin2019bert}, which are widely expected to be robust against domain shift because their representations are learned from the data alone, without a task label to over-specialize on.
We test this expectation in combination with domain adaptation.
The waveform encoder is trained first, on its own, with standard domain adaptation and no energy labels.
Having never seen the spectrum, it cannot encode it, and the pulse-shape difference is the only thing its critic can act on.
The encoder is then frozen, and the rest of the network is trained on top of it to reconstruct the energy.
We call this the \emph{staged domain adaptation} approach.

The second approach keeps the network end-to-end but localizes the critic to the level at which the mismatch actually lives.
Here the critic is attached to the output of the waveform encoder and acts on each station separately, with the ADA weight of an event propagated to all of its stations.
Section~\ref{sec:waveform} found a station critic to be harmful, but the situation there was different.
The waveform shift was accompanied by an LDF mismatch, which is not visible in a single station and thus fools it to wrong domain-invariance.
Here the accompanying mismatch is the spectrum, which ADA has already removed from the critic's view, and the pulse shape that remains is a genuine nuisance.
We further refer to this approach as waveform ADA, and to the end-to-end scheme with an event-wise critic as event ADA.

\begin{figure}[t]
\centering
\begin{subfigure}[b]{0.45\textwidth}
\centering
\includegraphics[width=\linewidth]{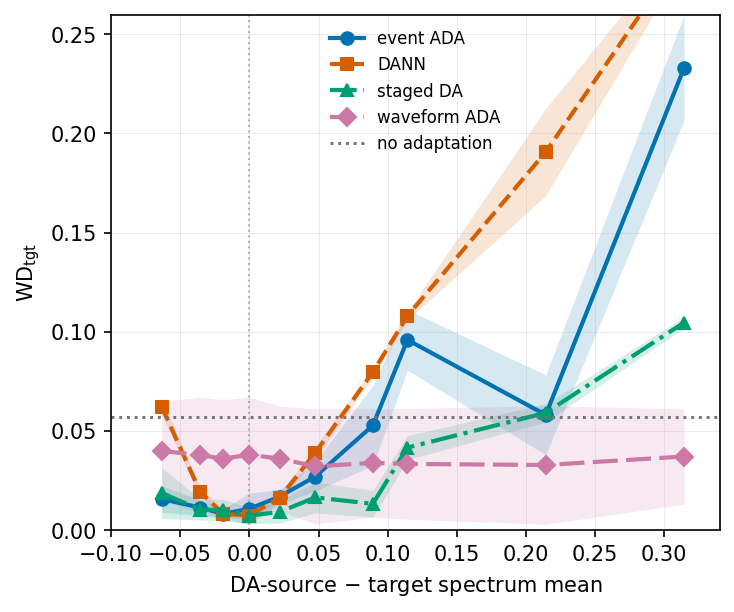}
\caption{}\label{fig:wfspec_wd}
\end{subfigure}
\hfill
\begin{subfigure}[b]{0.45\textwidth}
\centering
\includegraphics[width=\linewidth]{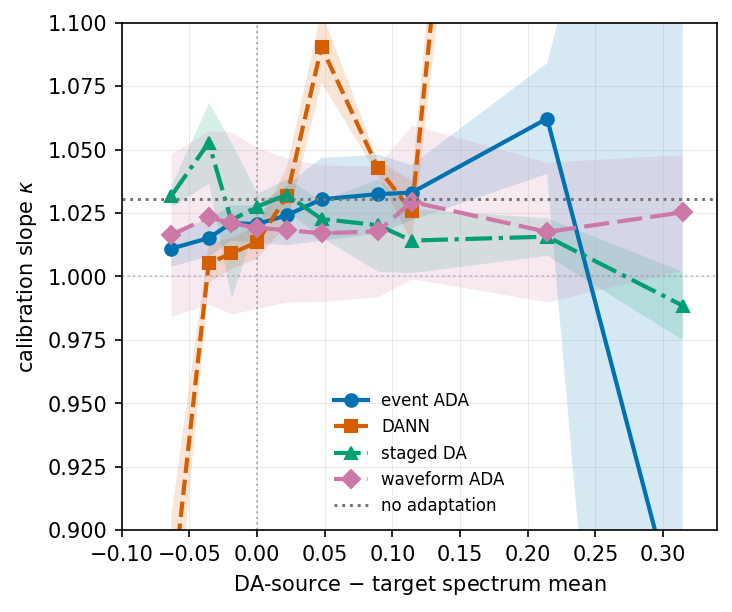}
\caption{}\label{fig:wfspec_k}
\end{subfigure}
\caption{Waveform-plus-spectrum mismatch for event ADA, DANN, staged DA, and waveform ADA with the no-adaptation baseline, against the difference between the DA-source and target spectrum means: \textbf{(a)} the target recovery $\mathrm{WD}_\mathrm{tgt}$ and \textbf{(b)} the calibration slope $\kappa$.}
\label{fig:wfspec}
\end{figure}

The results are presented in figure~\ref{fig:wfspec}.
The no-adaptation baseline reaches $\mathrm{WD}_\mathrm{tgt} = 0.057$.
Near matched spectra, event ADA and staged DA perform best, reaching $\mathrm{WD}_\mathrm{tgt} \approx 0.01$, well below the no-DA baseline.
Their error grows as the spectrum difference widens, reaching $\mathrm{WD}_\mathrm{tgt} = 0.10$--$0.23$ at a difference of $0.3$.
Waveform ADA behaves oppositely: it stays near $\mathrm{WD}_\mathrm{tgt} = 0.035$ across the whole range, but pays a higher floor at small spectrum shifts, where the other two are better.
The calibration slope stays close to unity, $\kappa \approx 1.0$--$1.05$, for all three throughout, so none of them distorts the shape of the spectrum.
Standard DANN, shown for reference, matches these methods only near coincident spectra and, as in the other compound cases, loses reliability as the spectra diverge.

Staged DA metrics become worse at large mismatch for a definite reason: the waveform encoder is de-nuisanced against a source whose spectrum is already wrong, and that bias passes into the readout trained on top of it.
This suggests the fix.
A trained model already estimates the target spectrum better than the guess it started from, so one can feed that estimate back as the DA-source prior and repeat.
The loop is then iterated until the feed-in and final energy spectra agree. 
We call this \emph{iterative domain adaptation}. 
For staged DA the refined prior enters at the first stage, where the waveform encoder is re-trained against a source whose spectrum matches the target more closely.

\begin{figure}[t]
\centering
\begin{subfigure}[b]{0.45\textwidth}
\centering
\includegraphics[width=\linewidth]{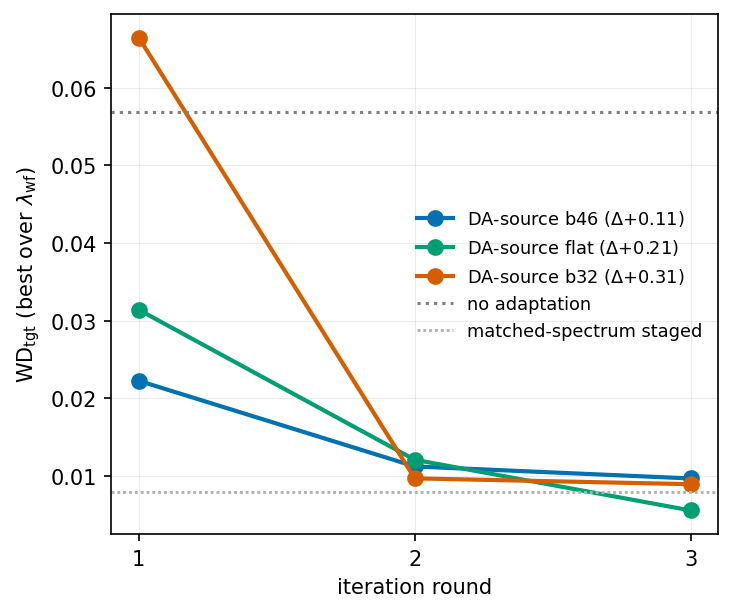}
\caption{}\label{fig:iterstaged_a}
\end{subfigure}
\hfill
\begin{subfigure}[b]{0.45\textwidth}
\centering
\includegraphics[width=\linewidth]{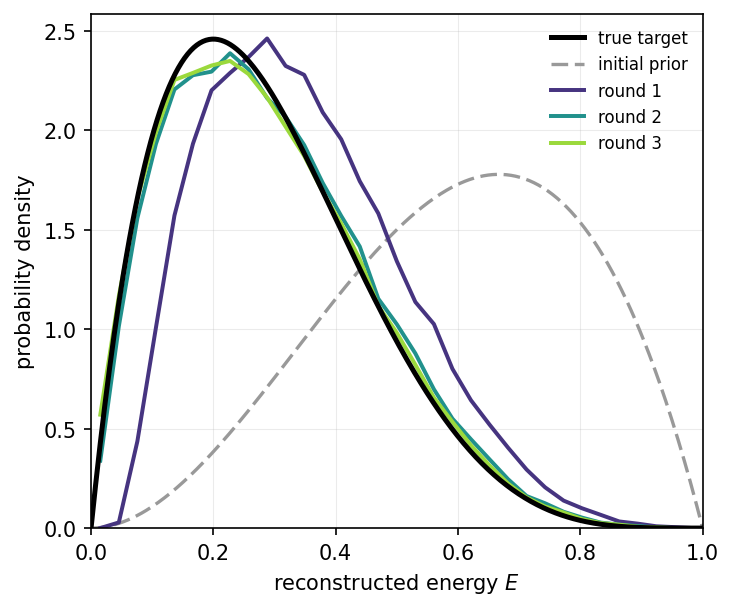}
\caption{}\label{fig:iterstaged_b}
\end{subfigure}
\caption{Iterative domain adaptation applied to staged DA. \textbf{(a)} Target recovery $\mathrm{WD}_\mathrm{tgt}$ against the iteration round, for three DA-source spectra at mean differences $+0.11$, $+0.21$ and $+0.31$. \textbf{(b)} Predicted target spectra across rounds for the largest mismatch, $\Delta=+0.31$.}
\label{fig:iterstaged}
\end{figure}

Figure~\ref{fig:iterstaged} shows the effect for three DA-source spectra spanning mean differences of $+0.11$ to $+0.31$. 
A single pass over-reads, the more so the larger the mismatch, but one or two rounds of refinement bring all three to the matched-spectrum floor of $\mathrm{WD}_\mathrm{tgt} \approx 0.01$, well below the no-DA baseline of $0.057$. 

With this refinement, staged DA is the most accurate method for a combined waveform-and-spectrum mismatch, reaching its matched-spectrum performance at any spectrum prior. 
Waveform ADA lags a little behind, holding $\mathrm{WD}_\mathrm{tgt} \approx 0.035$ across the whole range. 
Note that its metrics are flat across the whole mean-energy-difference range, with a higher floor than staged DA at the matched spectrum.
This implies that waveform DA does not benefit from iterative domain adaptation.

\subsection{All shifts at once: the realistic case}
\label{sec:triple}

We now consider the realistic case, in which all three domain shifts are present at once: the waveform mismatch (detector response), the energy-spectrum (label) shift, and the LDF slope (physical) mismatch.
As in the previous cases, we take the energy-dependent slope $\beta_\mathrm{tgt} = 3 \pm 0.2\,E$ and sweep the difference between the DA-source and target spectrum means. 
The target spectrum is $\mathrm{Beta}(2,5)$, the labelled source is flat, and the pulses are Gaussian in the source and Laplacian in the target.

In this case the two conditional shifts --- of the LDF slope and waveform shape --- are present and entangled.
As it was demonstrated in section~\ref{sec:waveform}, under such conditions a single critic performs better than two dedicated ones.
This logic also excludes staged DA as a viable option for the triple mismatch case.
Hence the only available option is ADA, which we compare against standard DANN.

\begin{figure}[t]
\centering
\begin{subfigure}[b]{0.45\textwidth}
\centering
\includegraphics[width=\linewidth]{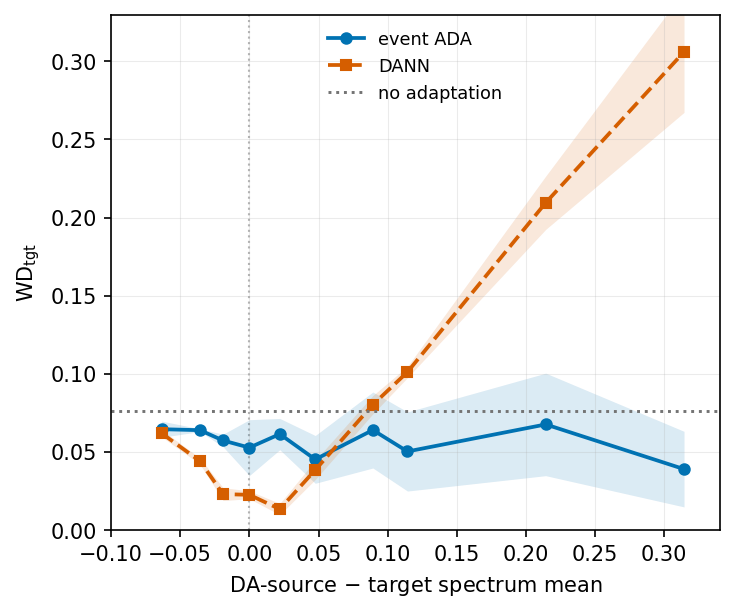}
\caption{}\label{fig:triple_a}
\end{subfigure}
\hfill
\begin{subfigure}[b]{0.45\textwidth}
\centering
\includegraphics[width=\linewidth]{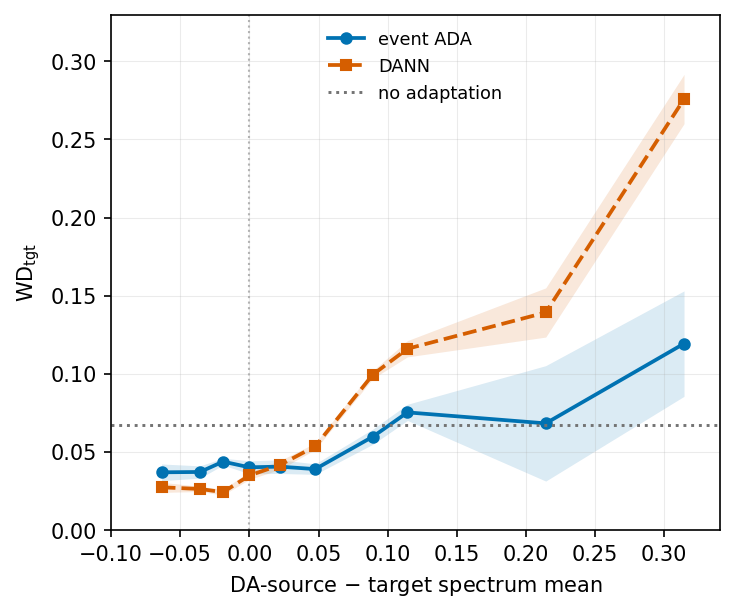}
\caption{}\label{fig:triple_b}
\end{subfigure}
\caption{Target recovery $\mathrm{WD}_\mathrm{tgt}$ against the DA-source---target spectrum-mean difference for event ADA and DANN for the triple mismatch at LDF slope \textbf{(a)} $+0.2E$ and \textbf{(b)} $-0.2E$.}
\label{fig:triple}
\end{figure}

The results are presented in figure~\ref{fig:triple} for different DA-source---target spectrum-mean differences at both slopes, with $\mathrm{MAE}_\mathrm{tgt}$ following the same trend.
The no-adaptation baseline lies at $\mathrm{WD}_\mathrm{tgt} = 0.08$ ($+0.2E$) and $0.07$ ($-0.2E$).
The two methods, ADA and DANN, agree only near matched spectra, where both recover the target and DANN is marginally better.
As the assumed DA-source spectrum departs from the target, DANN degrades rapidly --- at the largest mismatch its error reaches several times the baseline --- because it anchors the target predictions on that assumed spectrum.
ADA is more robust, changing only slowly across the sweep, which makes it the safer choice whenever the spectrum prior is imperfect.
The reversed slope is the hardest setting, where the over-reading bias established in section~\ref{sec:combined} compounds the spectrum mismatch: there even ADA rises to $\mathrm{WD}_\mathrm{tgt} \approx 0.12$ at the largest mismatch.

\begin{figure}[t]
\centering
\begin{subfigure}[b]{0.45\textwidth}
\centering
\includegraphics[width=\linewidth]{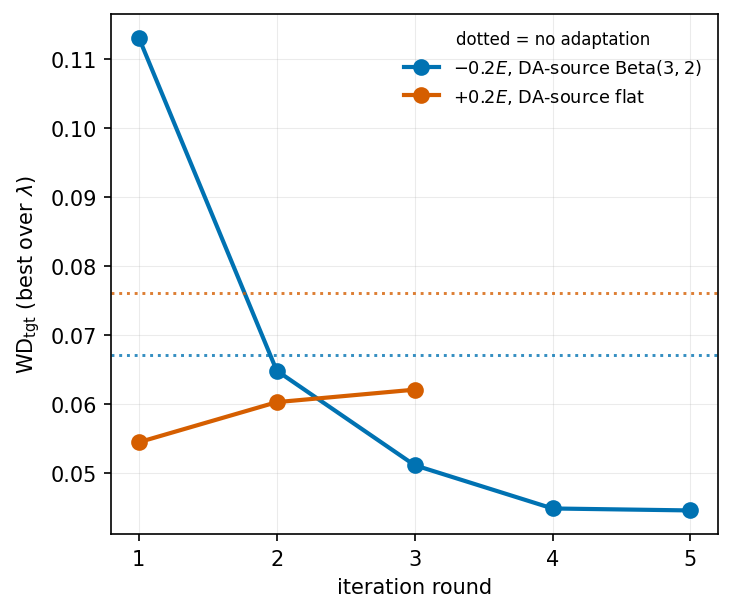}
\caption{}\label{fig:iterada_a}
\end{subfigure}
\hfill
\begin{subfigure}[b]{0.45\textwidth}
\centering
\includegraphics[width=\linewidth]{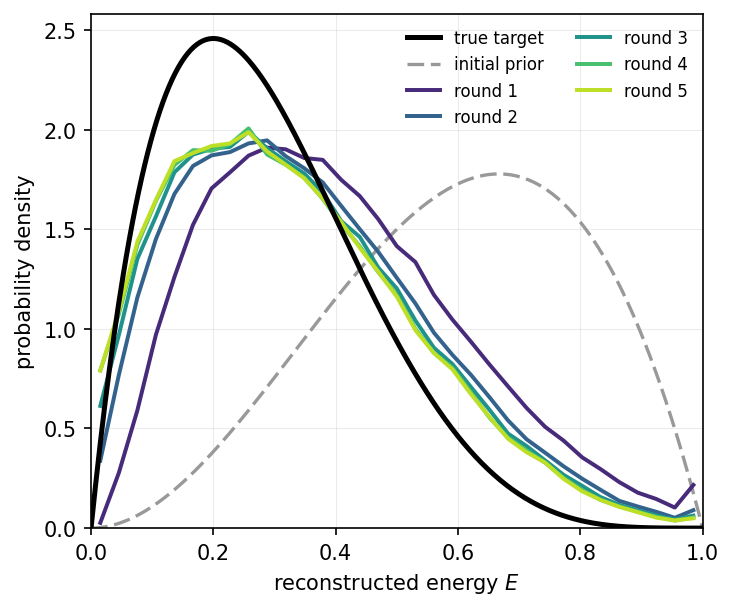}
\caption{}\label{fig:iterada_b}
\end{subfigure}
\caption{Iterative ADA under the triple mismatch. \textbf{(a)} Target recovery $\mathrm{WD}_\mathrm{tgt}$ against the iteration round, for a badly-mismatched prior ($\alpha=-0.2$, DA-source $\mathrm{Beta}(3,2)$) and a mild one ($\alpha=+0.2$, flat DA-source); round~1 is single-pass ADA and the dotted lines are the corresponding no-adaptation baselines. \textbf{(b)} Predicted target spectra across rounds for the $-0.2E$ case, converging from the initial prior towards the true $\mathrm{Beta}(2,5)$.}
\label{fig:iterada}
\end{figure}

The single-pass comparison above assumes a fixed DA-source spectrum, but the iterative domain adaptation of section~\ref{sec:wfspec} carries over to ADA: we feed each run's predicted target spectrum back as the DA-source prior and repeat. 
Figure~\ref{fig:iterada} shows the outcome at both slopes, and the gain is largest where the mismatch is largest. 
For the badly-chosen $\mathrm{Beta}(3,2)$ DA-source at LDF slope $-0.2E$, a single pass reaches only $\mathrm{WD}_\mathrm{tgt} = 0.113$, yet a few rounds drive it down to $0.045$.

Iterative ADA behaves differently for positive and negative LDF slope corrections.
At $+0.2E$ with a flat DA-source, where a single pass already reads the spectrum well, feeding its slightly under-read prediction back as the prior mildly reinforces the under-reading.
As a result, the error slightly drifts up across rounds, but only from $\mathrm{WD}_\mathrm{tgt} = 0.055$ to $0.062$.
This small cost buys a robustness check that needs no labels: one iterates and traces the predicted spectrum from round to round.
A spectrum that barely moves, as at $+0.2E$, confirms that the single pass was already stable and iterations can be stopped.
The spectrum that moves substantially, as at $-0.2E$, shows that the single pass was prior-limited and that the iterates are the better estimate.
The natural stopping rule is therefore to iterate until the predicted spectrum shifts only slightly or stops changing at all between rounds.

The results of this section demonstrate that even the triple domain mismatch case can be mitigated.
The key techniques for that are adaptive and iterative domain adaptations.

\section{Model selection without target labels}
\label{sec:stop}

Applying domain adaptation in practice requires somehow fixing the adversarial strength $\lambda$.
In a toy experiment one can pick the value that best recovers the known target labels, but in a real analysis they are unavailable by construction: the target energy spectrum is precisely the quantity to be measured, so the recovery $\mathrm{WD}_\mathrm{tgt}$ is unavailable.
A usable procedure therefore needs a rule that selects the operating point $\lambda$ from the training dynamics alone.

The rule we arrive at has three steps.
The first step --- whether to use domain adaptation at all --- is decided by the source-only reconstruction test of section~\ref{sec:spec}.
If the autoencoder trained on the source domain successfully reconstructs target events, that is, if the reconstruction loss is small, domain adaptation is not needed, and vice versa.

The next step is a \emph{regression-health gate}.
If the reconstruction loss function on the labelled training source --- energy reconstruction on $\mathcal{D}_\mathrm{tr}$ in our case --- exceeds a small multiple of the no-DA baseline, the adversarial pressure has collapsed the regressor.
We set the threshold $\mathrm{MAE}_\mathrm{src} > 3\,\mathrm{MAE}_\mathrm{src}^\mathrm{base}$ and discard all model configuration that loses a large fraction ($>30\%$) of its seeds this way.
The factor of three in $\mathrm{MAE}$ criterion is not a tuned quantity --- it has only to be loose enough to pass a healthy run and tight enough to catch a collapse.

Among the health-gated survivors we need a label-free signal that ranks the true recovery, and four candidates suggest themselves, each reading a different facet of the trained model.
\emph{Marginal separability} $\mathrm{LR}_\mathrm{acc}$ is the accuracy of a fresh logistic-regression domain classifier fit to the frozen event-summary vectors --- how far apart the two domains sit in latent space, ignoring the energy.
The \emph{conditional discrepancy} is the conditional maximum mean discrepancy $\mathrm{CMMD}^2$ of appendix~\ref{app:ssl}, which measures the same latent mismatch, but separately in bins of the predicted energy and averaged over them.
\emph{Augmentation consistency} is the stability of a run's target predictions under label-preserving perturbations of the input.
The \emph{cross-seed spread} is the mean pairwise Wasserstein distance between the predicted target spectra of independently seeded runs of one configuration.
The first two read the latent geometry; the last two read the stability of the predictions.

To choose between them, we screen the four on the waveform-and-LDF case of section~\ref{sec:waveform}.
Each candidate is correlated (Spearman $\rho$) against the true target recovery $\mathrm{WD}_\mathrm{tgt}$, oriented so that a positive value means the signal ranks recovery correctly, table~\ref{tab:selcrit}.
The purpose of this screen is not to establish the rule but to motivate it, by discarding the candidates that already fail on a case where domain adaptation works well.

\begin{table}[t]
\centering
\small
\begin{tabular}{l l c}
\toprule
candidate label-free signal & kind & $\rho$ \\
\midrule
marginal separability $\mathrm{LR}_\mathrm{acc}$ & marginal latent      & $+0.3$ \\
conditional discrepancy (CMMD)                    & conditional latent   & $+0.77$ \\
augmentation consistency                          & prediction stability & $-0.4$ to $-0.7$ \\
\textbf{cross-seed spread}                        & prediction stability & $\mathbf{+1.00}$ \\
\bottomrule
\end{tabular}
\caption{Label-free selection signals ranked (Spearman $\rho$) against the true target recovery $\mathrm{WD}_\mathrm{tgt}$ on the waveform-and-LDF case of section~\ref{sec:waveform}, oriented so that a positive value means the signal ranks recovery correctly.
The no-DA baseline is excluded from the ranking.}
\label{tab:selcrit}
\end{table}

Three of the four candidates fail, and they fail for two distinct reasons.
The two latent signals do not pass due to the physics of the problem.
Marginal separability was already compromised in section~\ref{sec:cov}: a successful adaptation makes the domains \emph{more} separable in latent space, not less, because the encoder has to read the local LDF shape in order to recover the energy slope-tolerantly.

The conditional discrepancy scores much better, but carries a structural flaw.
Its bins are defined by the model's own predicted energy, so a residually pooled model sorts its compressed predictions into the wrong bins.
Comparing latents within such a bin then compares events of genuinely different energies, and the model reports a small discrepancy for the wrong reason.
The no-DA baseline is the extreme case: it has a healthy source regression, the worst target recovery of all, and yet the lowest $\mathrm{CMMD}^2$ of any configuration.
We have also verified that this criterion does not fit more complicated cases and hence was discarded. 

Augmentation consistency fails for the other reason, and its correlation is not merely weak but negative --- it ranks the recovery backwards.
The cause is that it reads the stability of a \emph{single} model, and a confidently biased model is perfectly stable.
A network that has collapsed onto a narrow, wrong spectrum will reproduce that spectrum under any label-preserving perturbation of its input, and so scores as the most trustworthy of all.
Stability of one model is thus not evidence of anything; it is equally consistent with convergence and with confident failure.

The cross-seed spread avoids both traps because it reads neither the latent geometry nor any single model's predictions, but the agreement between independently seeded runs of the same configuration.
A bias can be stable within one training, but two trainings that start from different weights agree on a spectrum only if the training has actually converged to it.
The adversarial instability that produces a poor mean recovery is the same instability that scatters the seeds, so the label-free spread and the label-requiring recovery move together.
We measure it as the mean pairwise Wasserstein distance between the per-seed predicted target spectra, and pick the survivor with the smallest.

In our experiments we traced how the cross-seed spread depends on the $\lambda$ value.
In the limit $\lambda \to 0$ it vanishes continuously since a critic too weak to move the encoder leaves every seed at the no-DA baseline.
Thus, it is near zero on the under-adapted branch. 
As $\lambda$ increases, the cross seed spread rises to a maximum where the alignment first engages and independent seeds scatter.
It then falls to a local minimum at the operating point, where the seeds re-converge on the recovered spectrum.
At still larger $\lambda$ it grows again as the training destabilises.
The quantity to select is that \emph{interior} minimum --- the smallest spread among the $\lambda$ values lying above the engagement maximum. 
This is our third step, constituting the core of the label-free model-selection criterion. 
Applied in this form, the rule returns for every case studied above the configuration whose $\mathrm{WD}_\mathrm{tgt}$ is within 25\% of the best achievable result. 

In appendix \ref{app:ssl} we briefly discuss other domain adaptation schemes that we tried but which bring no success.

\section{Conclusion}
\label{sec:summ}
 
Domain shift between the training and target domains is one of the main obstacles to applying neural networks for data analysis in physics and beyond.
Without domain adaptation the estimated target properties --- the energy spectrum, in our case --- can be strongly biased.
Using a toy air-shower benchmark that exposes three distinct mismatches --- a waveform nuisance, a lateral-distribution physics shift, and an energy-spectrum label shift --- we showed that standard adversarial domain adaptation is not a safe default.
It handles a conditional shift on its own, but once a label shift is present and the simulated spectrum is imperfect, it pulls the predicted spectrum toward that simulation prior, biasing the very quantity to be measured.
A targeted method is needed to keep this bias bounded.
 
To resolve this problem, we introduce Adaptive Domain Adaptation (ADA), aimed at the label shift.
After each training epoch it reweights the DA-source events so that their predicted spectrum matches the target's.
This removes the energy spectrum from the domain-discriminative features and lets the adversary act only on the genuine physical shift.
ADA extends importance-weighted adversarial alignment from classification to label-free regression, and recovers a faithful target spectrum in the joint spectrum-and-physics regime where standard domain adaptation fails.
 
When the target spectrum is unknown, the assumed spectrum can itself be refined: the predicted target spectrum of one training round becomes the DA-source prior of the next, and the estimate converges to a self-consistent spectrum that approximates the truth.
This iterative domain adaptation corrects the assumed prior not only from epoch to epoch within a training but from round to round across trainings, and applies to any of the adaptation schemes above whenever the initial spectrum is only an approximation.

The predicted spectrum depends on the strength of domain adaptation, controlled by the $\lambda$ parameter.
To select its proper value we developed a label-free model-selection rule based on cross-seed spread of neural network predictions.
For properly converged domain adaptation this spread is minimal, indicating that the neural network learns the same features independently of the initial random state.
We verified that the rule returns a configuration within $25\%$ of the label-chosen optimum across every case studied, so that the methods above can be deployed where, by construction, no reliable target labels exist.

An interesting finding is that successful adaptation does not make the two domains indistinguishable in latent space.
Under the LDF mismatch a logistic regression fitted to the frozen event summaries separates them \emph{better} after adaptation than before.
The encoder has to read the local shape of the LDF in order to recover the energy slope-tolerantly, so the latents necessarily carry the slope, and with it the domain.
What adaptation makes invariant is the prediction, not the representation, and diagnostics built on latent domain-invariance are therefore reading the wrong quantity.
 
Table~\ref{tab:guidance} summarises which adaptation scheme to use depending on the domain mismatch type.
When the source-only reconstruction test of section~\ref{sec:spec} finds no shift, the network already interpolates correctly and no adaptation should be applied.
When a shift is present and the assumed spectrum is trustworthy, standard DANN is the method of choice: sections~\ref{sec:combined} and~\ref{sec:triple} show it to be slightly better in that narrow band.
When the assumed spectrum is only an approximation, the choice turns on whether the domain shift can be localized.
If it can --- as with the waveform mismatch of section~\ref{sec:wfspec} --- staged DA, with its source spectrum refined by iteration, is the most accurate, and a localized ADA critic is a simpler alternative not far behind.
If it cannot, the event-wise ADA critic is the safer choice, since sub-critics see only a part of the mismatch and are biased by the part they cannot resolve.
In either case iterative domain adaptation --- feeding the network's own prediction back as the prior --- refines the assumed spectrum and brings the estimate closer to the truth.

\begin{table}[t]
\centering
\small
\begin{tabularx}{\textwidth}{@{}l l X@{}}
\toprule
knowledge of the target spectrum & domain shift & recommended scheme \\
\midrule
--- & none detected & no adaptation; the network already interpolates \\
trustworthy & any & DANN with an event-wise critic \\
approximate & localized & staged DA or ADA with localized critic \\
approximate & not localized & ADA with an event-wise critic \\
\bottomrule
\end{tabularx}
\caption{Practical guidance on the choice of the adaptation scheme.}
\label{tab:guidance}
\end{table}

The discussed mismatch regimes are often encountered in physics. 
They arise whenever a network is trained on simulations known to be imperfect and applied to data whose distribution is itself the quantity being measured.
Such a situation is rather common in astroparticle physics.
Cosmic-ray and neutrino observatories reconstruct events with neural networks~\cite{Kalashev:2021vop, Glombitza:2020yhw, Huennefeld:2017pdh, Kharuk:baikal}, and estimating the energy spectrum is among their primary physics goals~\cite{kim2023energy, PierreAuger:2021tog}.
The muon puzzle, however, remains a simulation-to-data discrepancy that needs to be mitigated to obtain a reliable spectrum estimate.
Moreover, the complexity of the detector system, simplifications in MC simulations, and approximations of the particle cross-section at high energies inevitably lead to domain shift.
Hence domain adaptation is a prerequisite for reliable data analyses.

Cosmology offers another promising application of these methods.
In this field neural networks are trained on simulations, labels for the observed data are unavailable, and the distribution of the inferred quantity is itself the measurement.
In photometric redshift estimation this distribution is calibrated on a spectroscopic sample that is not representative of the survey.
In simulation-based inference~\cite{Swierc:2024gdu, Pandya:2025zak} it is fixed by the cosmology assumed when generating the training data.
In both cases a standard adversary would pull the result towards that prior, just as it pulls the reconstructed spectrum towards the simulated one above.
With surveys such as Rubin and Euclid now delivering unlabelled data on a large scale~\cite{LSST:2008ijt, Euclid:2025rvk}, these methods might find a useful application.

\section*{Acknowledgments}
This work was supported by the Russian Science Foundation under grant no. 24-72-10056.

\appendix
 
\section{Methods that did not work}
\label{app:ssl}
 
In order to improve domain adaptation, we tried two other methods, based on self-supervised learning (SSL) and maximum mean discrepancy (MMD).
None of them, however, helped, and below we briefly report on them.
 
Following the logic of foundation models~\cite{devlin2019bert}, we tried implementing SSL.
The network carries an auxiliary decoder and is trained to reconstruct the full event --- the per-station waveforms and scalars --- from the event summary, on both source and target events.
The intent is to stop the encoder from folding target events onto the source manifold: a representation that must reconstruct genuine target events cannot discard everything that makes them different, which should limit the pooling toward the DA-source physics.
 
In practice SSL hurt the adaptation, raising $\mathrm{WD}_\mathrm{tgt}$ by $10$--$30\%$ across the cases we tried.
The reconstruction objective teaches the network what a target event looks like, but supplies no energy label, so it does not help the regression.
Added on top of the already adversarial domain-adaptation loss, it only makes the shared encoder's optimization harder and degrades the recovered spectrum.
This is the same lesson as the staged encoder of section~\ref{sec:wfspec}: a representation learned without labels is robust against the mismatch it was trained to ignore, but does not help to restore true labels.
 
The second method is the maximum mean discrepancy (MMD)~\cite{gretton2012kernel}, in its conditional form.
Given two batches of latent vectors $\{z^\mathrm{src}_i\}_{i=1}^{N_s}$ and $\{z^\mathrm{tgt}_j\}_{j=1}^{N_t}$ and a kernel $k(z, z')$, the empirical squared MMD is
\begin{equation}
\label{eq:mmd}
\mathrm{MMD}^2 \;=\;
\frac{1}{N_s^2}\sum_{i,j} k(z^\mathrm{src}_i, z^\mathrm{src}_j)
+ \frac{1}{N_t^2}\sum_{i,j} k(z^\mathrm{tgt}_i, z^\mathrm{tgt}_j)
- \frac{2}{N_s N_t}\sum_{i,j} k(z^\mathrm{src}_i, z^\mathrm{tgt}_j) \;.
\end{equation}
It vanishes in the large-sample limit if and only if the two batches are drawn from the same underlying latent distribution.
We use a Gaussian kernel $k(z,z') = \exp(-\gamma \|z-z'\|^2)$ summed over a short multi-bandwidth set $\gamma \in \{0.1, 1, 10\}$, so the result is insensitive to per-feature scaling and detects differences of any order beyond a linear classifier's reach.
 
For a regression one wants latents to agree at \emph{fixed label}, since that is what determines whether a source-trained regressor gives the right answer on a target event.
Target labels are unavailable, but the network's predicted energy $\hat E$ is a deployable proxy.
Binning source and target events by $\hat E$ into $B$ uniform bins on $[0,1]$, computing $\mathrm{MMD}^2_b$ inside each bin and averaging over bins gives the \emph{conditional} maximum mean discrepancy:
\begin{equation}
\label{eq:cmmd}
\mathrm{CMMD}^2 \;=\; \frac{1}{B} \sum_{b} \mathrm{MMD}^2_b \;.
\end{equation}
We use $B = 10$ to test this approach and verified that the results are qualitatively the same for $B$ between $5$ and $20$. The CMMD loss is added to the total loss function with a tunable coefficient.
 
This construction, however, has the same weakness that disqualified $\mathrm{CMMD}^2$ as a selection signal in section~\ref{sec:stop}: the bins are set by the model's own predicted energy.
A confidently biased model sorts its shifted target predictions into the wrong bins and then forces the latents of genuinely different-energy source and target events into alignment.
The result is a self-reinforcing loop --- the alignment deepens the bias, which mis-bins the events further --- because the criterion has no internal handle on the true target energy.
Combining CMMD with ADA did not remove this failure: CMMD keeps over-aligning the wrong latent subspaces, and at no setting of its coefficient did the combination improve on ADA alone.

\bibliography{refs}

\end{document}